\documentclass[10pt,twocolumn,letterpaper]{article}

\usepackage[pagenumbers]{cvpr}

\usepackage{graphicx}
\usepackage{amsmath}
\usepackage{amssymb}
\usepackage{booktabs}
\usepackage{bigstrut}
\usepackage{multirow}
\usepackage{array}
\usepackage{tabularx}
\usepackage{tabu}
\usepackage{amsmath}
\usepackage{amsthm}
\usepackage[pagebackref,breaklinks,colorlinks]{hyperref}

\usepackage[capitalize]{cleveref}
\crefname{section}{Sec.}{Secs.}
\Crefname{section}{Section}{Sections}
\Crefname{table}{Table}{Tables}
\crefname{table}{Tab.}{Tabs.}

\def\cvprPaperID{7055} % *** Enter the CVPR Paper ID here
\def\confName{CVPR}
\def\confYear{2022}

\begin{document}

\title{Colar: Effective and Efficient Online Action Detection by Consulting Exemplars}
\author{Le\ Yang \ \ \ \ Junwei\ Han\thanks{Corresponding author.} \ \ \ \ Dingwen\ Zhang \\School of Automation, \ \ Northwestern Polytechnical University, \ \ China \\ \href{https://nwpu-brainlab.gitee.io/index_en}{https://nwpu-brainlab.gitee.io/index\_en} }

\maketitle

%%%%%%%%% ABSTRACT
\begin{abstract}

Online action detection has attracted increasing research interests in recent years. Current works model historical dependencies and anticipate the future to perceive the action evolution within a video segment and improve the detection accuracy. However, the existing paradigm ignores category-level modeling and does not pay sufficient attention to efficiency. Considering a category, its representative frames exhibit various characteristics. Thus, the category-level modeling can provide complimentary guidance to the temporal dependencies modeling. This paper develops an effective exemplar-consultation mechanism that first measures the similarity between a frame and exemplary frames, and then aggregates exemplary features based on the similarity weights. This is also an efficient mechanism, as both similarity measurement and feature aggregation require limited computations. Based on the exemplar-consultation mechanism, the long-term dependencies can be captured by regarding historical frames as exemplars, while the category-level modeling can be achieved by regarding representative frames from a category as exemplars. Due to the complementarity from the category-level modeling, our method employs a lightweight architecture but achieves new high performance on three benchmarks. In addition, using a spatio-temporal network to tackle video frames, our method makes a good trade-off between effectiveness and efficiency. Code is available at \href{https://github.com/VividLe/Online-Action-Detection}{https://github.com/VividLe/Online-Action-Detection.}

\end{abstract}

\section{Introduction}

\begin{figure}[htbp]
\centering
\includegraphics[width=1\linewidth]{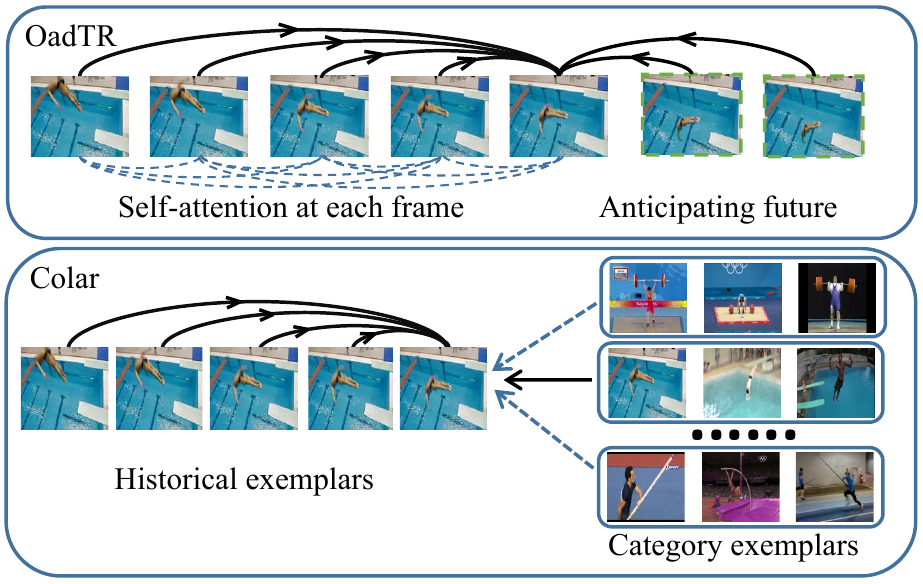}
\caption{Comparison between existing state-of-the-art method OadTR \cite{wang2021oadtr} and our proposed Colar. Unlike OadTR, Colar consults historical exemplars to model long-term dependencies and consults category exemplars to capture category-level particularity, forming an effective and efficient method.}
\label{figMotivation}
\vspace{-0.4cm}
\end{figure}

With the development of mobile communications, video has become a powerful medium to record life and transform information. As a result, video understanding technologies have aroused increasing research interests. Among these technologies, temporal action detection \cite{shou2016temporal, wang2021rgb, zhu2021enriching} can discover action instances from untrimmed videos and extract valuable information. Well-performed action detection algorithms can benefit smart surveillance \cite{mhalla2018embedded}, anomaly detection \cite{chandola2009anomaly} \etc In recent years, along with action detection technologies becoming mature, a more challenging but more practical task, namely online action detection, has been proposed \cite{de2016online}. The online action detection algorithm tackles a streaming video, reports the occurrence of an action instance, and keeps alarming until the action ends \cite{de2016online}. In inference, the algorithm only employs historical frames that have been observed, but has no access to future frames.

As an early exploration, Geest~\etal \cite{de2016online} discovered the importance of modeling long-term dependencies. Later, Xu \etal \cite{xu2019temporal} revealed the value of anticipating future status to enhance the long-term dependencies modeling. OadTR \cite{wang2021oadtr} recently utilized the multi-head self-attention module to jointly model historical dependencies and anticipate the future, which achieved promising online action detection results.

As an under-explored domain, there are three core challenges for online action detection: How to model long-term dependencies? How to associate a frame with representative frames from the same category? How to conduct detection efficiently? Existing works \cite{de2016online, xu2019temporal, wang2021oadtr} primarily focus on the long-term dependencies modeling, but ignore the other two challenges. However, as shown in Figure \ref{figMotivation} (a), both analyzing historical frames and anticipating future status only model relationships within a video segment, leaving the category-level modeling under-explored. Because an action category contains multiple instances and each instance exhibits special appearance and motion characteristic, the guidance of exemplary frames can make the online detection algorithm more robust to resist noises within a video segment. In addition, a practical online action detection algorithm should always consider the computational efficiency, including both the efficiency to perform online detection and the efficiency to extract video features.

This paper develops an exemplar-consultation mechanism to tackle above three challenges in a unified framework. The exemplar-consultation mechanism first jointly transforms a frame and its exemplary frames to the key space and value space. Then, it measures the similarity in the key space and employs the similarity to aggregate information in the value space. As both feature transformation and similarity measurement require limited computations, the proposed exemplar-consultation mechanism is efficient. Considering a video segment, we can effectively model long-term dependencies by using historical frames as exemplars based on the exemplar-consultation mechanism. As we only compare one frame with its historical frames, rather than performing self-attention on all frames, the computational burden is alleviated. Similarly, we can also regard representative frames of each category as exemplars and conduct category-level modeling based on the exemplar-consultation mechanism. Compared with a video segment, category exemplars can provide complementary guidances and make the algorithm more robust.

\begin{figure*}[thbp]
\centering
\includegraphics[width=1\linewidth]{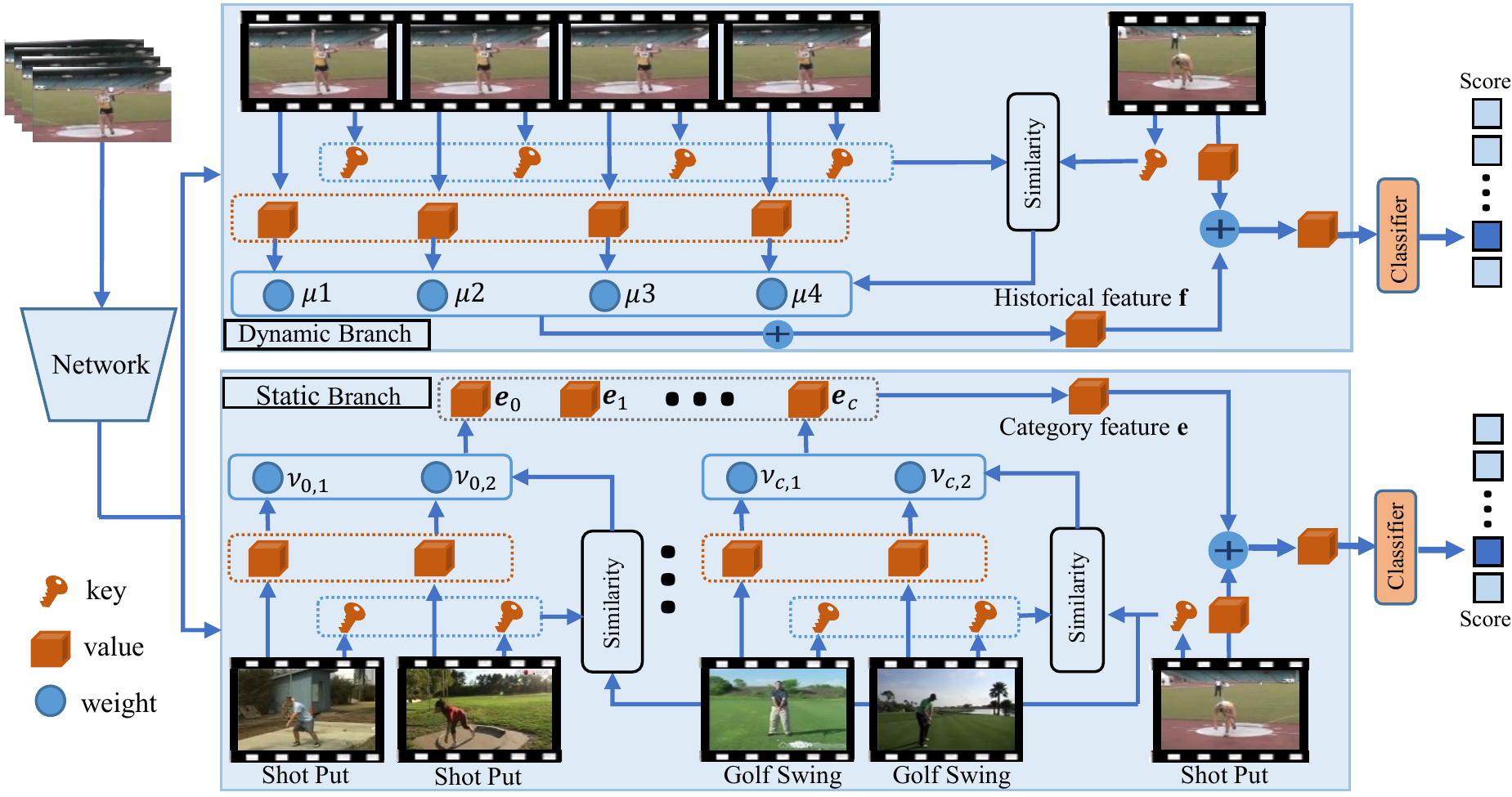}
\caption{Framework of the proposed Colar method for online action detection. Given a video, the dynamic branch compares a frame with its historical exemplars and models temporal dependencies, while the static branch compares a frame with category exemplars and captures the category particularity.}
\label{figFramework}
\end{figure*}

By \underline{\textbf{co}}nsulting exemp\underline{\textbf{lar}}s, we build a unified framework, namely Colar, to perform online action detection, as shown in Figure \ref{figFramework}. Colar maintains the dynamic branch and the static branch in parallel, where the former models long-term dependencies within a video segment and the latter models category-level characteristics. In the dynamic branch, Colar consults previous frames and aggregates historical features. In the static branch, Colar first obtains category exemplars via clustering, then consults exemplars and aggregate category features. Finally, two classification scores are fused to detect actions. Moreover, we analyze the running time bottleneck of existing works and discover the expensive costs to extract flow features. Thus, we employ a spatio-temporal network to only dispose of video frames and perform end-to-end online action detection, which only takes 9.8 seconds to tackle a one-minute video. To sum up, this paper makes the following contributions:
\begin{itemize}
    \item We make an early attempt to conduct category-level modeling for the online action detection task, which provides holistic guidance and makes the detection algorithm more robust.
    \item We propose the exemplar-consultation mechanism to compare similarities and aggregate information, which can efficiently model long-term dependencies and category particularities.
    \item Due to the effectiveness of the exemplar-consultation mechanism and the complimentary guidance from category-level modeling, our method employs a lightweight architecture. Still, it achieves superior performance and builds new state-of-the-art performance on three benchmarks.
\end{itemize}

\section{Related work}

\noindent
\textbf{Modeling temporal dependencies.} Different from image-based task, \eg detection \cite{feng2020progressive, feng2020tcanet, yao2020automatic, feng2021saenet}, localization\cite{zhu2021few, guo2021strengthen, zhang2021weakly} and segmentation\cite{zhangD2021weakly}, it is crucial to model temporal dependencies for online action detection. Existing works rely on recurrent networks, including both LSTM-based methods \cite{de2018modeling, yuan2017temporal, xu2019temporal} and GRU-based methods \cite{eun2020learning}. Specifically, Geest~\etal \cite{de2018modeling} proposed a two-stream LSTM \cite{hochreiter1997long} network. Similarly, TRN \cite{xu2019temporal} employed LSTM blocks to model historical temporal dependencies. Recently, OadTR~\cite{wang2021oadtr} drove the recurrent-network paradigm into a transformer-based paradigm and effectively captured the long-term relationship via self-attention. Although OadTR~\cite{wang2021oadtr} effectively models long-term dependencies, the self-attention process for all frames leads to the computational burden problem. This work regards historical frames as exemplars and utilizes the exemplar-consultation mechanism to model long-term dependencies.

\noindent
\textbf{Anticipating future.} Although online action detection algorithms cannot access future frames, anticipating future features can assist the decision of current frame. In RED \cite{gao2017red}, Gao \etal estimated features for future frames and calculated the classification loss and the feature regression loss to improve the anticipation quality, which is further developed by TRN \cite{xu2019temporal} and OadTR \cite{wang2021oadtr}. In this paper, the static branch employs the exemplar-consultation mechanism to compare a frame with representative exemplars of each category and brings complementary information to the dynamic branch.

\noindent
\textbf{Offline action detection.}
The offline action detection algorithm aims to discover action instances from untrimmed videos \cite{chen2019relation, liu2021multi, liu2021end}, where all video frames can be utilized. Some algorithms~\cite{shou2016temporal, xu2017r, wang2021rgb, huang2021clrnet} tackled video frames to perform localization. In addition, a majority of works \cite{lin2017single, zhu2021enriching} first extracted video features from powerful backbone networks\cite{wang2016temporal, carreira2017quo, zhu2020label, zhu2021temporal}, then performed action localization based on video features. From the view of anchor mechanism, the representative works include anchor-based methods \cite{lin2017single, zhu2021enriching} and anchor-free methods \cite{lin2019bmn, yang2020revisiting, lin2021learning}. Besides, multiple effective modules have been proposed, \eg graph convolutional module \cite{zeng2019graph, zeng2021graph}. Moreover, action detection under the weakly supervised setting \cite{zeng2019breaking, yang2021background, zhao2021soda, yuan2019marginalized} was also well explored.

The primary difference between online action detection and offline action detection algorithms lies in whether future frames can be accessed. In offline algorithms, Xu \etal \cite{xu2017r} performed data augmentation via playing the video in reverse order, while Zhu \cite{zhu2021enriching} modeled the relationship among multiple proposals within a video. However, these procedures are unsuitable for the studied online action detection task. 

\noindent
\textbf{Space-time memory network.}
Oh \etal \cite{oh2019video} proposed the space-time memory network to efficiently connect a frame and its previous frames via space-time memory read. It has verified effective performance on modeling temporal information, and has been extended to multiple tasks,~\eg video object detection \cite{chen2020memory}, video object segmentation \cite{lu2020video, huang2021scribble}, tracking \cite{lai2020mast}. In contrast to the space-time memory network, our static branch employs the exemplar-consultation mechanism to model the intra-category relationship. Specifically, it first aggregates particular features from each category, and then combines multiple features to obtain the category feature.

\section{Method}

Given a video stream, the online action detection algorithm should report the occurrence once the action starts and keep alarming until the action ends. The learning process is guided by the frame-level classification label $\mathbf{y}=[y_{0}, y_{1}, ..., y_{C}]$, where $y_{c} \in \{0, 1\}$ indicates whether frame $\mathbf{f}_{0}$ belongs to the $c^{th}$ category. As shown in Figure \ref{figFramework}, we first employ a backbone network to extract video features. Then, we propose the dynamic branch to model long-term dependencies within a segment and propose the static branch to capture the holistic particularity for each category. Finally, two detection results are fused to perform the online action detection task.

\subsection{Dynamic branch}

As neighboring frames can provide rich contextual cues to determine the category label of the current frame, the core idea of the dynamic branch is to model local evolution by comparing a frame with its previous historical frames and dynamically aggregating the local features. The upper part of Figure \ref{figFramework} exhibits the detailed operations in the dynamic branch. Compared with the standard multi-head self-attention mechanism of OadTR\cite{wang2021oadtr}, our proposed dynamic branch makes two reasonable designs, which sufficiently benefit the online action detection task. First, we use temporal convolution with kernel size 3 to model local cues among historical frames, which is complementary to the global modeling of self-attention. Second, we make two simplifications over OadTR, \ie removing the class token and replacing multi-head self-attention with one-head attention on the current frame. The simplifications reduce learning difficulty and benefit the performance when training data is not rich enough.

Given a video feature sequence, we first transform a feature $\mathbf{f}_{t}$ to the key space and the value space, where the former is responsible for comparing similarity, and the latter can be used for feature aggregation.
\begin{equation}
    \mathbf{f}_{t}^{k} = \Phi^{k}(\mathbf{f}_{t}), \ \ \mathbf{f}_{t}^{v} = \Phi^{v}(\mathbf{f}_{t}),
\end{equation}
where $\Phi^{k}$ and $\Phi^{v}$ indicate two convolutional layers in the dynamic branch. Then, we measure the pair-wise affinity between $\mathbf{f}_{0}^{k}$ and other key features (\eg $\mathbf{f}_{t}^{k}$) via calculating the cosine similarity:
\begin{equation}
    \mu_{t} = cos(\mathbf{f}_{0}^{k}, \mathbf{f}_{t}^{k}) = \frac{\mathbf{f}_{0}^{k} \cdot \mathbf{f}_{t}^{k}}{||\mathbf{f}_{0}^{k}|| \cdot ||\mathbf{f}_{t}^{k}||}.
\end{equation}

Given a series of affinity values $[\mu_{-T}, ..., \mu_{-1}, \mu_{0}]$, we perform softmax normalization and obtain the attention mask $[\hat{\mu}_{-T}, ..., \hat{\mu}_{-1}, \hat{\mu}_{0}]$. As each element $\mu_{t}$ indicates the similarity between the previous $t^{th}$ frame and the current frame, we can aggregate value features among previous frames and obtain the historical feature $\mathbf{f}$:
\begin{equation}
    \mathbf{f} = \sum_{t=-T}^{0} \hat{\mu}_{t} \cdot \mathbf{f}_{t}^{v}.
\end{equation}
In the end, the dynamic branch jointly considers value feature $\mathbf{f}_{0}^{v}$ and the historical feature $\mathbf{f}$ (\eg via summation) and conduct online action detection:
\begin{equation}
    \mathbf{s}^{\rm d} = \Omega^{\rm d}(\mathbf{f}_{0}^{v}, \mathbf{f} | \Theta^{\rm d}),
\end{equation}
where $\Omega^{\rm d}$ is the classifier in the dynamic branch with parameter $\Theta^{\rm d}$, and $\mathbf{s}^{\rm d} \in \mathbb{R}^{C+1}$ is the classification score from the dynamic branch.

\subsection{Static branch}

Considering action instances from the same category, some instances with distinctive appearance characteristics and clear motion patterns can be selected as exemplars to represent this category. We employ the K-means clustering algorithm for each category, carry out clustering, and obtain $M$ exemplary features. On this basis, the online action detection task can be formulated as comparing a frame with representative exemplars of each category. As a result, the static branch can provide complementary cues to the dynamic branch and makes the online detection algorithm robust to noises within the local video segment.

Before stepping to detailed operations in the static branch, it is necessary to analyze its efficiency. First, using another branch increases a certain computation. However, compared with OadTR\cite{wang2021oadtr}, we not only simplify the attention computation but also remove decoder layers. Thus, our holistic computation is smaller than OadTR\cite{wang2021oadtr}, and we require less memory as well (see experiments in Sec.\ref{ExpCmp}). In addition, even given a dataset with millions of samples and thousands of categories, the modern implementation\cite{johnson2019billion} of the K-Means algorithm can still efficiently generate exemplars, as verified by DeepCluster\cite{caron2018deep}.

As shown in the bottom part of Figure \ref{figFramework}, the static branch operates with the category exemplars $\{\mathcal{E}_{c} = [\mathbf{e}_{c, 1}, \mathbf{e}_{c, 2}, ..., \mathbf{e}_{c, M}]\}_{c=0}^{C}$ to classify feature $\mathbf{f}_{0}$, where each category contains $M$ representative exemplars. At first, we convert each exemplar $\mathbf{e}_{c,i}$ to the key space and the value space:
\begin{equation}
    \mathbf{e}_{c,i}^{k} = \Psi^{k}(\mathbf{e}_{c,i}), \ \ \mathbf{e}_{c,i}^{v} = \Psi^{v}(\mathbf{e}_{c,i}),
\end{equation}
and convert the frame feature $\mathbf{f}_{0}$ to the key space and value space as well:
\begin{equation}
    \mathbf{e}_{0}^{k} = \Gamma^{k}(\mathbf{f}_{0}), \ \ \mathbf{e}_{0}^{v} = \Gamma^{v}(\mathbf{f}_{0}),
\end{equation}
where $\Psi^{k}$, $\Psi^{v}$, $\Gamma^{k}$ and $\Gamma^{v}$ indicate convolutional layers. In the key space, we can measure the similarity between feature $\mathbf{f}_{0}$ and exemplar $\mathcal{E}_{c}$ from the $c^{th}$ category:
\begin{equation}
    \nu_{c,i} = cos(\mathbf{e}_{0}^{k}, \mathbf{e}_{c,i}^{k}) = \frac{\mathbf{e}_{0}^{k} \cdot \mathbf{e}_{c,i}^{k}}{\|\mathbf{e}_{0}^{k}\| \cdot \|\mathbf{e}_{c,i}^{k}\|}.
\end{equation}
Based on the pair-wise similarity between $\mathbf{e}_{0}^{k}$ and all exemplars $[\mathbf{e}_{c, 1}, \mathbf{e}_{c, 2}, ..., \mathbf{e}_{c, M}]$ from the $c^{th}$ category, we can first calculate the attention mask $[\hat{\nu}_{c, 1}, ..., \hat{\nu}_{c, M}]$ via softmax normalization, and then aggregate all exemplars to represent the current frame from the perspective of the $c^{th}$ category:
\begin{equation}
    \mathbf{e}_{c} = \sum_{i=1}^{M} \hat{\nu}_{c, i} \cdot \mathbf{e}_{c,i}^{v}.
\end{equation}

After comparing the current frame with representative exemplars of all categories, we obtain category-specific features $[\mathbf{e}_{0}, \mathbf{e}_{1}, ..., \mathbf{e}_{C}]$. Considering a feature from the $c^{th}$ category, it would be similar to exemplars from the $c^{th}$ category while be different from other exemplars. Thus, we use a convolutional layer to estimate the attention weight $\mathbf{a} \in \mathbb{R}^{C+1}$ and aggregate category feature $\mathbf{e}$:
\begin{equation}
    \mathbf{e} = \sum_{c=0}^{C} a_{c} \cdot \mathbf{e}_{c}.
\end{equation}

The exemplary feature $\mathbf{e}$ is generated from all exemplars and can reveal the category characteristics. In the end, the static branch employs both value feature $\mathbf{e}_{0}^{v}$ and category feature $\mathbf{e}$ to predict the classification score $\mathbf{s}^{\rm s}$:
\begin{equation}
    \mathbf{s}^{\rm s} = \Omega^{\rm s}(\mathbf{e}_{0}^{v}, \mathbf{e} | \Theta^{\rm s}),
\end{equation}
where $\Omega^{\rm s}$ is the classifier with parameter $\Theta^{\rm s}$.

\subsection{Efficient online action detection}
Given a series of pre-extracted video features, the dynamic branch connects a frame with its historical neighbors and models local evolution, while the static branch compares a frame with representative exemplars and models category particularity. It is convenient to fuse the predictions of two branches and detect actions online. However, the feature extraction process, especially calculating optical flows, requires heavy computations, which prevents us from conducting online action detection in practical scenarios.

To alleviate the computational burdens, we can employ a spatio-temporal network to tackle video frames and provide representative features for the dynamic and the static branch. Considering the video recognition performance and calculation efficiency, we utilize the ResNet-I3D network \cite{wang2018non}, discard the last classification layer, and construct our feature extraction backbone. Given a video sequence with $T$ frames, the output of the backbone network is $\mathbf{x} \in \mathbb{R}^{D \times T/8}$, where $D$ indicates the feature dimension. In practice, as the benchmark datasets contain limited training videos, we find frozen the first three blocks can produce more accurate detection results.

\subsection{Training and inference}

Given a frame, the dynamic branch and the static branch predict its classification score $\mathbf{s}^{\rm d}$ and $\mathbf{s}^{\rm s}$, respectively. We calculate the cross-entropy loss to guide the learning process:
\begin{equation}
    \mathcal{L}_{cls}^{\rm d} = -\sum_{c=0}^{C} \mathbf{y}_{c} {\rm log}(\hat{\mathbf{s}_{c}}^{\rm d}), \ \ \mathcal{L}_{cls}^{\rm s} = -\sum_{c=0}^{C} \mathbf{y}_{c} {\rm log}(\hat{\mathbf{s}_{c}}^{\rm s}),
\end{equation}
where $\hat{\mathbf{s}_{c}}^{\rm d}$ and $\hat{\mathbf{s}_{c}}^{\rm s}$ indicate scores after softmax normalization. Besides, as the dynamic and static branches tackle the same frame, two classification scores should be consistent. Thus, we introduce the consistency loss $\mathcal{L}_{cons}$ to enable mutual guidance among two branches:
\begin{equation}
    \mathcal{L}_{cons} = \mathcal{L}_{KL}(\hat{\mathbf{s}}^{\rm d} \| \hat{\mathbf{s}}^{\rm s}) + \mathcal{L}_{KL}(\hat{\mathbf{s}}^{\rm s} \| \hat{\mathbf{s}}^{\rm d}),
\end{equation}
where $\mathcal{L}_{KL}$ indicates the KL-divergence loss. As verified by Zhang \etal \cite{zhang2018deep}, the consistency loss can lead to a robust model with better generalization. To sum up, the training process is guided by the following loss:
\begin{equation}
    \mathcal{L} = \mathcal{L}_{cls}^{\rm d} + \mathcal{L}_{cls}^{\rm s} + \lambda \mathcal{L}_{cons},
\end{equation}
where $\lambda$ is a trade-off parameter.

\begin{table}[thbp]
  \centering
  \caption{Comparison experiments on THUMOS14 dataset, measured by mAP (\%).}
    \begin{tabular}{cc|c}
    \toprule
    \toprule
    Setups & Method & mAP(\%) \\
    \midrule
    \midrule
          & CNN\cite{simonyan2014very}{\scriptsize ICLR15} & 34.7 \\
          & CNN\cite{simonyan2014two}{\scriptsize NIPS14} & 36.2 \\
    Offline & LRCN\cite{donahue2015long}{\scriptsize CVPR15} & 39.3 \\
          & MultiLSTM\cite{yeung2018every}{\scriptsize IJCV18} & 41.3 \\
          & CDC\cite{shou2017cdc}{\scriptsize CVPR17} & 44.4 \\
    \midrule
          & RED\cite{gao2017red}{\scriptsize BMVC17} & 45.3 \\
          & TRN\cite{xu2019temporal}{\scriptsize ICCV19} & 47.2 \\
    Online & IDU\cite{eun2020learning}{\scriptsize CVPR20} & 50.0 \\
    (TSN-Anet) & OadTR\cite{wang2021oadtr}{\scriptsize ICCV21} & 58.3 \\
          & Colar & \textbf{59.4} \\
    \midrule
          & IDU\cite{eun2020learning}{\scriptsize CVPR20} & 60.3 \\
    Online & OadTR\cite{wang2021oadtr}{\scriptsize ICCV21} & 65.2 \\
    (TSN-Kinetics) & Colar & \textbf{66.9} \\
    \midrule
    RGB end-to-end & Colar & 58.6 \\
    \bottomrule
    \end{tabular}%
  \label{tabCmpTHUMOS}%
\end{table}%

In inference, the dynamic classification score $\mathbf{s}^{\rm d}$ and the static classification score $\mathbf{s}^{\rm s}$ are fused via a balance coefficient $\beta$ to perform online action detection:
\begin{equation}
    \mathbf{s} = \beta\ \hat{\mathbf{s}}^{\rm s} + (1 - \beta)\ \hat{\mathbf{s}}^{\rm d}.
    \label{eqFusion}
\end{equation}

\section{Experiments}

\subsection{Setups}

\textbf{Dataset.} We carry out experiments on three widely used benchmarks, THUMOS14 \cite{THUMOS14}, TVSeries \cite{de2016online} and HDD \cite{ramanishka2018toward}. THUMOS14 \cite{THUMOS14} includes sports videos from 20 action categories, where the validation set and test set contain 200 and 213 videos, respectively. On THUMOS14, challenges for online action detection include drastic intra-category varieties, motion blur, short action instances \etc We follow previous works \cite{de2018modeling, xu2019temporal, eun2020learning, wang2021oadtr}, train the model on the validation set, and evaluate performance on the test set.

TVSeries~\cite{de2016online} collects about 16 hours of videos from 6 popular TV series. The dataset contains 30 daily actions, where the total instance number is 6231. The TVSeries dataset exhibits some challenging characteristics, \eg temporal overlapping action instances, a large proportion of background frames and unconstrained perspectives.

HDD \cite{ramanishka2018toward} contains 104 hours of human driving video, belonging to 11 action categories. The videos were collected from 137 driving sessions using an instrumented vehicle equipped with different sensors. Following existing works \cite{de2018modeling, xu2019temporal, wang2021oadtr}, we use 100 sessions for training and 37 sessions for testing.

\begin{table}[thbp]
  \centering
  \caption{Comparison experiments on TVSeries dataset, measured by mcAP(\%).}
  \begin{tabular}{cc|c}
    \toprule
    \toprule
    Setups & Method & mcAP(\%) \\
    \midrule
    \midrule
          & LRCN\cite{donahue2015long}{\scriptsize CVPR15} & 64.1 \\
          & RED\cite{gao2017red}{\scriptsize BMVC17} & 71.2 \\
    RGB   & 2S-FN\cite{de2018modeling}{\scriptsize WACV18} & 72.4 \\
          & TRN\cite{xu2019temporal}{\scriptsize ICCV19} & 75.4 \\
          & IDU\cite{eun2020learning}{\scriptsize CVPR20} & 76.6 \\
    \midrule
    \multirow{2}[2]{*}{Flow} & FV-SVM\cite{de2016online}{\scriptsize ECCV2016} & 74.3 \\
          & IDU\cite{eun2020learning}{\scriptsize CVPR20} & 80.3 \\
    \midrule
          & RED\cite{gao2017red}{\scriptsize BMVC17} & 79.2 \\
          & TRN\cite{xu2019temporal}{\scriptsize ICCV19} & 83.7 \\
    Online & IDU\cite{eun2020learning}{\scriptsize CVPR20} & 84.7 \\
    (TSN-Anet) & OadTR\cite{wang2021oadtr}{\scriptsize ICCV21} & 85.4 \\
          & Colar & \textbf{86.0} \\
    \midrule
          & IDU\cite{eun2020learning}{\scriptsize CVPR20} & 86.1 \\
    Online & OadTR\cite{wang2021oadtr}{\scriptsize ICCV21} & 87.2 \\
    (TSN-Kinetics) & Colar & \textbf{88.1} \\
    \midrule
    RGB end-to-end & Colar & 86.8 \\
    \bottomrule
    \end{tabular}%
  \label{tabCmpTVSeries}%
\end{table}%

\textbf{Metric.} We adopt mean average precision (mAP) and calibrated mean average precision (cmAP) to measure the performance of online action detection algorithms. As for mAP, we first collect classification scores for all frames and then calculate precision and recall based on sorted results. Afterward, we calculate interpolated average precision to obtain AP scores for a category and finally regard the mean value of AP scores among all categories as mAP. Considering the drastically imbalanced frame numbers of different categories, Geest \etal \cite{de2016online} proposed to calibrate the mAP score. In particular, we first calculate the ratio $w$ between background frames and action frames and then calculate the calibrated precision as:
\begin{equation}
    cPre(i) = \frac{w \cdot TP(i)}{w \cdot TP(i) + FP(i)}.
\end{equation}
Afterward, the calibrated average precision $cAP$ for a category can be calculated as:
\begin{equation}
    cAP = \frac{\sum_{i} cPre(i) \cdot \mathbf{1}(i)}{\sum_{i} \mathbf{1}(i)},
\end{equation}
where $\mathbf{1}(\cdot)$ indicates whether the $i^{th}$ frame belongs to the considered action category. Finally, cmAP can be obtained via calculating the mean value among all cAPs.

\begin{table}[thbp]
  \centering
  \caption{Comparison experiments on HDD dataset, measured by mAP (\%).}
  \setlength{\tabcolsep}{8pt}
    \begin{tabular}{rc|c}
    \toprule
    \toprule
    \multicolumn{1}{l}{Setups} & \multicolumn{1}{l|}{Method} & \multicolumn{1}{l}{mAP(\%)} \\
    \midrule
    \midrule
          & CNN \cite{de2016online}{\scriptsize ICLR15} & 22.7 \\
          & LSTM \cite{ramanishka2018toward}{\scriptsize CVPR18} & 23.8 \\
    \multicolumn{1}{l}{Sensors} & ED \cite{gao2017red}{\scriptsize BMVC17} & 27.4 \\
          & TRN \cite{xu2019temporal}{\scriptsize ICCV19} & 29.2 \\
          & OadTR \cite{wang2021oadtr}{\scriptsize ICCV21} & 29.8 \\
          & Colar & \textbf{30.6} \\
    \bottomrule
    \end{tabular}%
  \label{tabCmpHDD}%
\end{table}%

\begin{table}[thbp]
  \centering
  \caption{Comparison between our proposed Colar method and existing methods. The inference time (in second) is measured on a 1080Ti GPU when tackling the same one-minute video. Both ``Colar*" and Colar$\dag$ directly tackle video frames, where the former uses a fixed backbone and the latter is end-to-end trained.}
  \setlength{\tabcolsep}{1.5pt}
  \small
    \begin{tabular}{c|cccc|c|c}
    \toprule
    \toprule
    \multirow{2}[2]{*}{Method} & RGB   & Optical & Flow  & Action & {\footnotesize Inference} & mAP \\
          & Feature & Flow  & Feature & Detection & Time  & (\%) \\
    \midrule
    \midrule
    \multicolumn{7}{c}{\footnotesize Given pre-extracted features, Colar is faster and more accurate.} \\
    \midrule
    IDU\cite{eun2020learning} & 2.3   & 39.8  & 4.4   & 52.8  & 99.3  & 60.3 \\
    OadTR\cite{wang2021oadtr} & 2.3   & 39.8  & 4.4   & 4.7   & 51.2  & 65.2 \\
    Colar & 2.3   & 39.8  & 4.4   & 4.2   & \textbf{50.7} & \textbf{66.9} \\
    \midrule
    OadTR-Flow & -     & 39.8  & 4.4   & 4.5   & 48.7  & 57.8 \\
    Colar-Flow & -     & 39.8  & 4.4   & 4.0   & \textbf{48.2} & \textbf{59.6} \\
    \midrule
    OadTR-RGB & 2.3   & -     & -     & 4.5   & 6.8   & 51.2 \\
    Colar-RGB & 2.3   & -     & -     & 4.0   & \textbf{6.3} & \textbf{52.1} \\
    \midrule
    \midrule
    \multicolumn{7}{c}{\footnotesize Given frames, Colar provides a trade-off between speed and accuracy.} \\
    \midrule
    Colar* & 5.8   & -     & -     & 4.0   & 9.8   & 53.4  \\
    Colar$\dag$ & 5.8   & -     & -     & 4.0   & 9.8   & \textbf{58.8}  \\
    \bottomrule
    \bottomrule
    \end{tabular}%
  \label{tabCmpRunTime}%
\end{table}%

\begin{table*}[thbp]
  \centering
  \caption{Detailed online action detection performances under different action portions, measured by mcAP (\%) on TVSeries dataset.}
  \setlength{\tabcolsep}{3.5pt}
   \begin{tabular}{cc|cccccccccc}
    \toprule
    \toprule
    \multirow{2}[2]{*}{Setups} & \multirow{2}[2]{*}{Method} & \multicolumn{10}{c}{Portion of actions} \\
          &       & 0-0.1 & 0.1-0.2 & 0.2-0.3 & 0.3-0.4 & 0.4-0.5 & 0.5-0.6 & 0.6-0.7 & 0.7-0.8 & 0.8-0.9 & 0.9-1 \\
    \midrule
    \midrule
    \multirow{2}[2]{*}{RGB} & CNN\cite{de2016online}{\scriptsize ICLR15} & 61.0  & 61.0  & 61.2  & 61.1  & 61.2  & 61.2  & 61.3  & 61.5  & 61.4  & 61.5 \\
          & LSTM\cite{de2016online}{\scriptsize ICLR15} & 63.3  & 64.5  & 64.5  & 64.3  & 65.0  & 64.7  & 64.4  & 64.4  & 64.4  & 64.3 \\
    \midrule
    Flow  & {\footnotesize FV-SVM}\cite{de2016online}{\scriptsize ECCV2016} & 67.0  & 68.4  & 69.9  & 71.3  & 73.0  & 74.0  & 75.0  & 75.4  & 76.5  & 76.8 \\
    \midrule
          & TRN\cite{xu2019temporal}{\scriptsize ICCV19} & 78.8  & 79.6  & 80.4  & 81.0  & 81.6  & 81.9  & 82.3  & 82.7  & 82.9  & 83.3 \\
    Online & IDU\cite{eun2020learning}{\scriptsize CVPR20} & \textbf{80.6} & 81.1  & 81.9  & 82.3  & 82.6  & 82.8  & 82.6  & 82.9  & 83.0  & 83.9 \\
    (TSN-Anet) & OadTR\cite{wang2021oadtr}{\scriptsize ICCV21} & 79.5  & 83.9  & 86.4  & 85.4  & 86.4  & 87.9  & 87.3  & 87.3  & 85.9  & 84.6 \\
          & Colar & 80.2  & \textbf{84.4} & \textbf{87.1} & \textbf{85.8} & \textbf{86.9} & \textbf{88.5} & \textbf{88.1} & \textbf{87.7} & \textbf{86.6} & \textbf{85.1} \\
    \midrule
          & IDU\cite{eun2020learning}{\scriptsize CVPR20} & 81.7  & 81.9  & 83.1  & 82.9  & 83.2  & 83.2  & 83.2  & 83.0  & 83.3  & 86.6 \\
    Online & OadTR\cite{wang2021oadtr}{\scriptsize ICCV21} & 81.2  & 84.9  & 87.4  & 87.7  & 88.2  & 89.9  & 88.9  & 88.8  & 87.6  & 86.7 \\
    (TSN-Kinetics) & Colar & \textbf{82.3} & \textbf{85.7} & \textbf{88.6} & \textbf{88.7} & \textbf{88.8} & \textbf{91.2} & \textbf{89.6} & \textbf{89.9} & \textbf{88.6} & \textbf{87.3} \\
    \midrule
    RGB end-to-end & Colar & 80.8  & 84.4  & 87.2  & 87.5  & 87.8  & 89.4  & 88.4  & 88.5  & 87.3  & 86.4 \\
    \bottomrule
    \end{tabular}%
  \label{tabCmpPortion}%
  \vspace{-0.3cm}
\end{table*}%

\textbf{Implementation details.} Following previous works \cite{de2018modeling, xu2019temporal, eun2020learning, wang2021oadtr}, we first conduct experiments with pre-extracted features. The feature extractor uses the two-stream network \cite{xiong2016cuhk}, whose spatial stream adopts ResNet-200 \cite{he2016deep} and temporal stream adopts BN-Inception \cite{ioffe2015batch}. We report two experiments where the two-stream network \cite{wang2016temporal, xiong2016cuhk} is trained on the ActivityNet v1.3 dataset \cite{caba2015activitynet} or the Kinetics-400 \cite{carreira2017quo} dataset to verify the generalization of the proposed Colar method. As for end-to-end online action detection, our backbone network is based on the ResNet50-I3D architecture \cite{wang2018non}, where the last average pooling layer and classification layer are removed. The ResNet50-I3D network is pretrained on Kinetics-400 \cite{carreira2017quo} dataset, and we use the weight file provided by MMAction2\cite{2020mmaction2}. In training, we freeze the first three blocks of the backbone network. Video frames are extracted with a frame rate of 25\textit{fps}, where the spatial size is set as 224$\times$224. We use the Adam \cite{kingma2014adam} algorithm to optimize the whole network and set the batchsize as 16. The initial learning rate is $3 \times 10^{-4}$ and decays every five epochs.

\subsection{Comparison experiments}
\label{ExpCmp}

\textbf{Quantitative comparisons.} We make a comparison with current state-of-the-art methods \cite{de2018modeling, xu2019temporal, eun2020learning, wang2021oadtr} and consistently build new high performance on THUMOS14 \cite{THUMOS14}, TVSeries \cite{de2016online}, and HDD \cite{ramanishka2018toward} benchmarks. As shown in Table \ref{tabCmpTHUMOS}, based on TSN-ActivityNet features, our Colar brings an mAP gain of 1.1\% over OadTR \cite{wang2021oadtr}, and the improvements would be an mAP of 1.7\% if the comparison is based on TSN-Kinetics features. The consistent improvements over current state-of-the-art methods verify the efficacy of our proposed exemplar-consultation mechanism. In addition, the proposed Colar can directly tackle video frames and perform online action detections, which achieves 58.6\% mAP. In addition to THUMOS14, experiments on TVSeries \cite{de2016online} and HDD \cite{ramanishka2018toward} benchmarks also verify the superiority of our method, as shown in Table \ref{tabCmpTVSeries} and Table \ref{tabCmpHDD}.

\textbf{Effectiveness and efficiency.} Table \ref{tabCmpRunTime} analyzes the performance and running time under different setups. When using pre-extracted features, Colar performs superior to existing methods \cite{eun2020learning, wang2021oadtr}. It is worth noting that extracting RGB and flow features takes 46.5 seconds, of which calculating optical flow costs the majority of the time. When only flow features or RGB features are available, the feature extraction cost is reduced, but both OadTR [34] and our Colar observe performance drop. In particular, it costs 44.2s to extract flow features, where OadTR [34] and Colar observe 7.4\% and 7.3\% performance drops, respectively. The cost to extract RGB features is small, but the online detection performance decreases a lot.

Given the ResNet50-I3D network \cite{wang2018non}, we first extract features from video frames and then train the proposed Colar method, which gets 53.4\%. In contrast, the proposed end-to-end learning paradigm achieves 58.8\%. To sum up, our proposed Colar method achieves a good balance between effectiveness and efficiency. Given pre-extracted features, Colar makes accurate detection results. Given only video frames, Colar costs 9.8 seconds to tackle a one-minute video and achieves comparable performance. In addition, we measure the memory cost under identical setups, where Colar requires 2235M memory and OadTR\cite{wang2021oadtr} requires 4375M memory.

\begin{table}[thbp]
  \centering
  \caption{Ablation studies about the efficacy of each component, measured by mAP(\%) on three benchmarks.}
  \setlength{\tabcolsep}{2.5pt}
    \begin{tabular}{ccc|ccc}
    \toprule
    \toprule
    Dynamic & Static & $\mathcal{L}_{cons}$ & THUMOS14 & TVSeries & HDD \\
    \midrule
    \midrule
    \checkmark &       &       & 65.2  & 86.3  & 29.5 \\
          & \checkmark &       & 58.1  & 83.5  & 26.4 \\
    \checkmark & \checkmark &       & 65.8  & 86.9  & 29.9 \\
    \checkmark & \checkmark & \checkmark & 66.9  & 88.1  & 30.6 \\
    \bottomrule
    \end{tabular}%
  \label{tabAblComplements}%
\end{table}%

\textbf{Performance under different action portions.} Table \ref{tabCmpPortion} elaborately studies the online action detection performance when different action portions are observed. The proposed Colar achieves promising accuracy when using the TSN-ActivityNet feature, the TSN-Kinetics feature, and only video frames. In particular, considering the most severe cases that only the first 10\% portion of actions are observed, the previous state-of-the-art method OadTR \cite{wang2021oadtr} shows inferior performance to IDU \cite{eun2020learning}. However, the proposed Colar consistently exceeds OadTR, due to that the static branch effectively connects a frame with representative exemplars of each category and provides complimentary guidance.

\begin{figure}[thbp]
     \centering
     \begin{subfigure}[b]{0.22\textwidth}
         \centering
         \includegraphics[width=\textwidth]{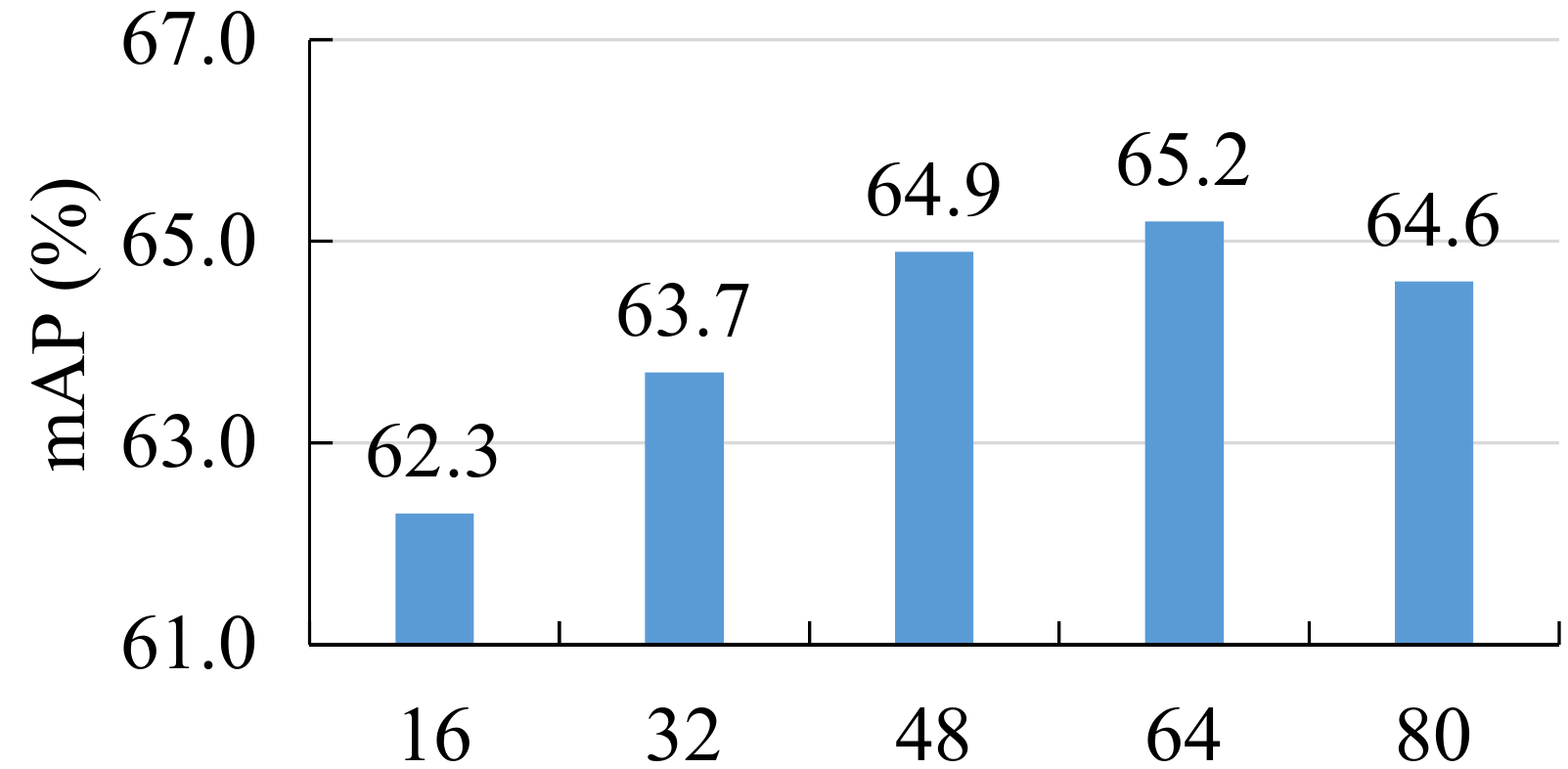}
         \caption{\footnotesize Temporal length $T$ in the dynamic branch.}
         \label{figAblTemLength}
     \end{subfigure}
     \begin{subfigure}[b]{0.22\textwidth}
         \centering
         \includegraphics[width=\textwidth]{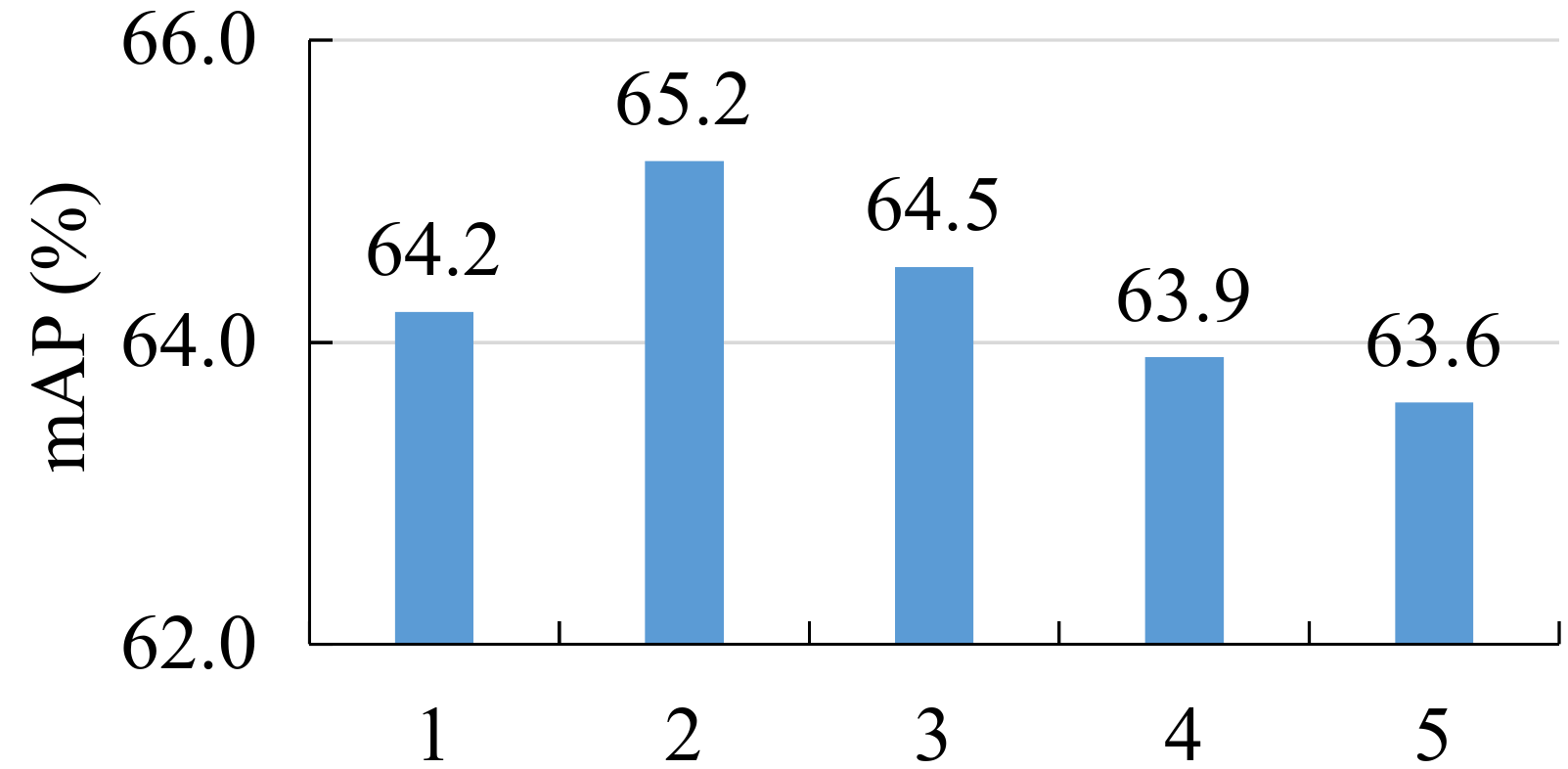}
         \caption{\footnotesize Number of convolutional layers in the dynamic branch.}
         \label{figAblConLayNum}
     \end{subfigure}
     \begin{subfigure}[b]{0.22\textwidth}
         \centering
         \includegraphics[width=\textwidth]{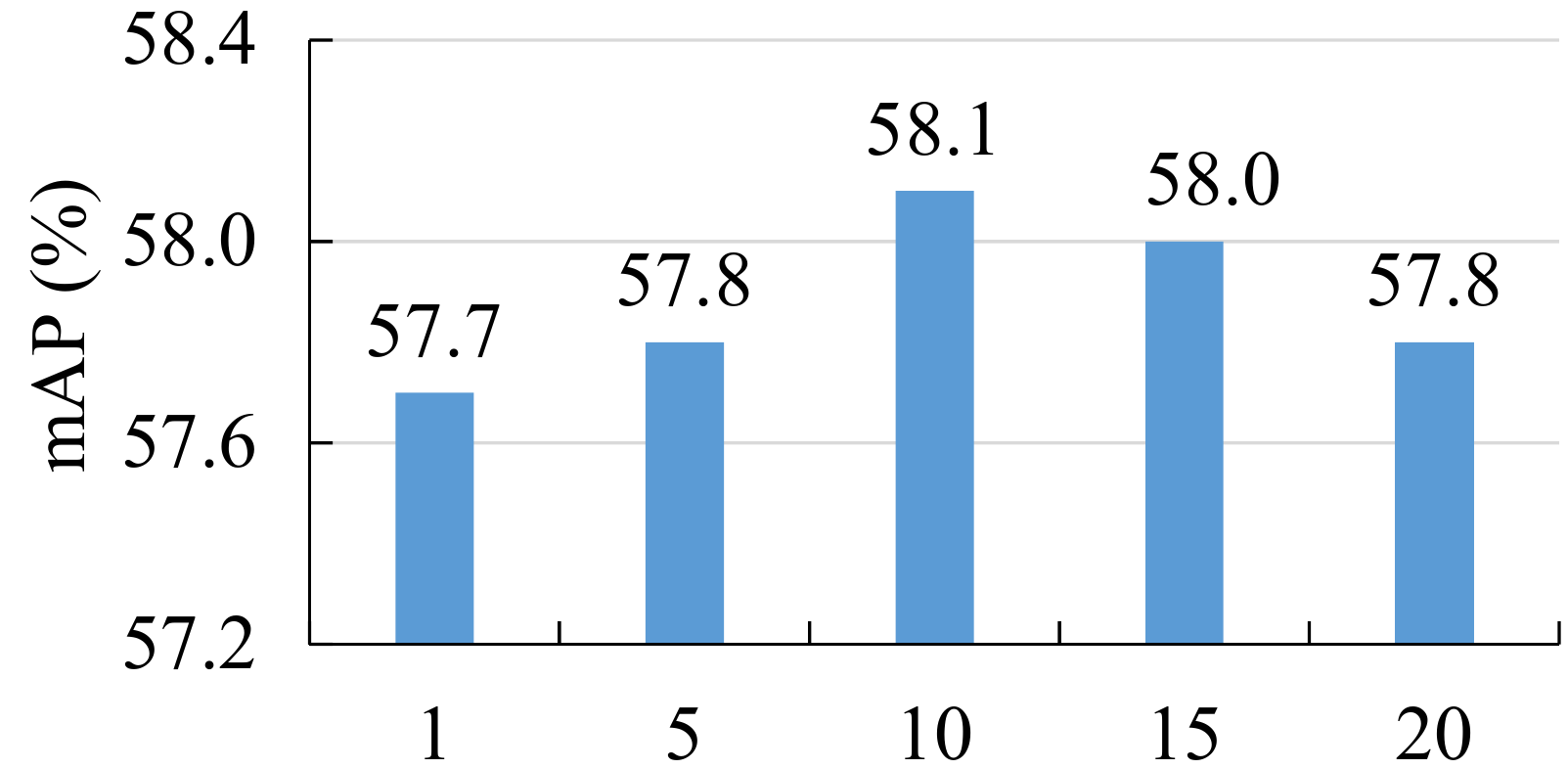}
         \caption{\footnotesize Number of representative features in the static branch.}
         \label{figAblFeatNum}
     \end{subfigure}
     \begin{subfigure}[b]{0.22\textwidth}
         \centering
         \includegraphics[width=\textwidth]{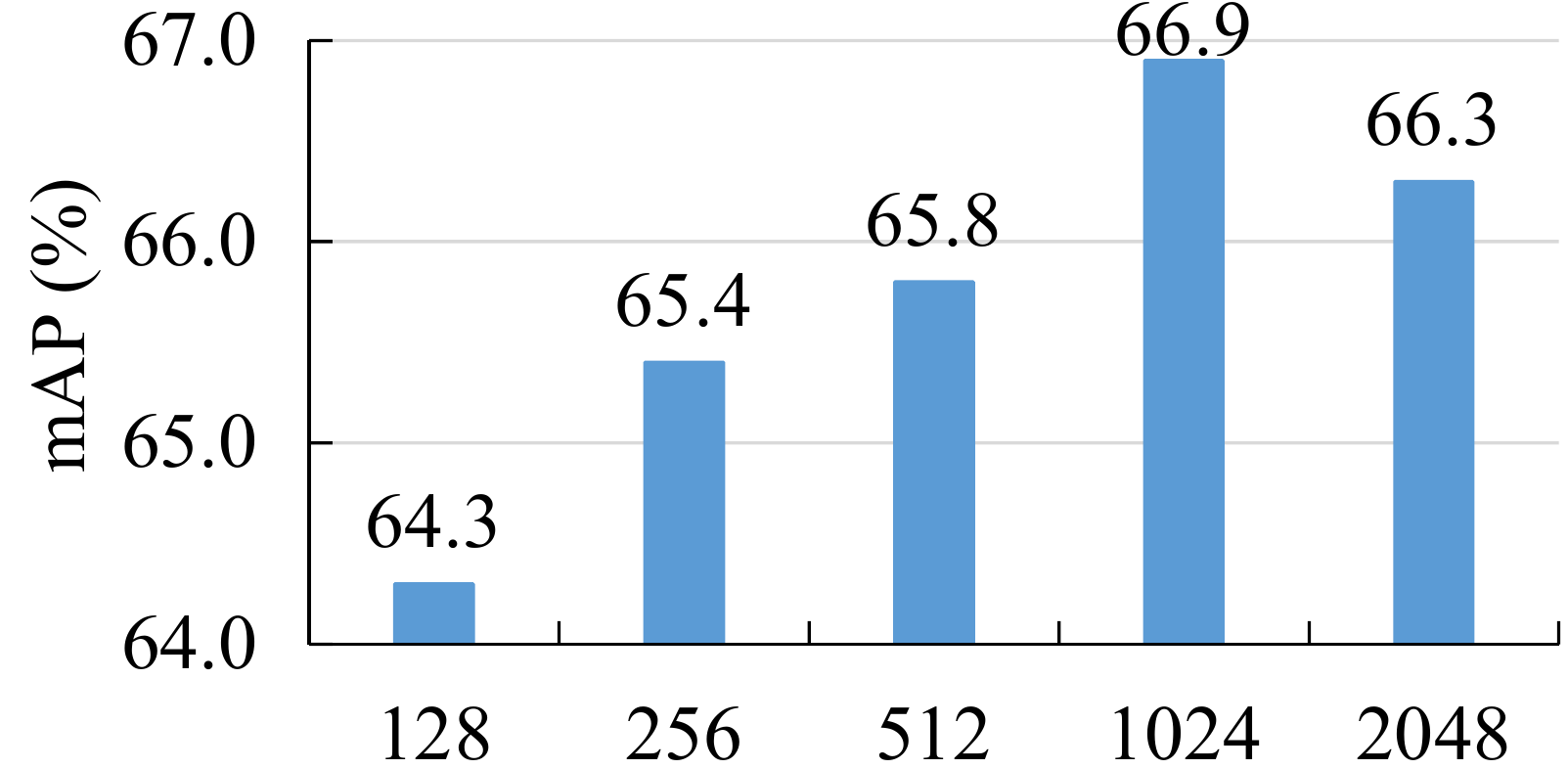}
         \caption{\footnotesize Channel number for the complete Colar method.}
         \label{figAblF}
     \end{subfigure}
     \begin{subfigure}[b]{0.22\textwidth}
         \centering
         \includegraphics[width=\textwidth]{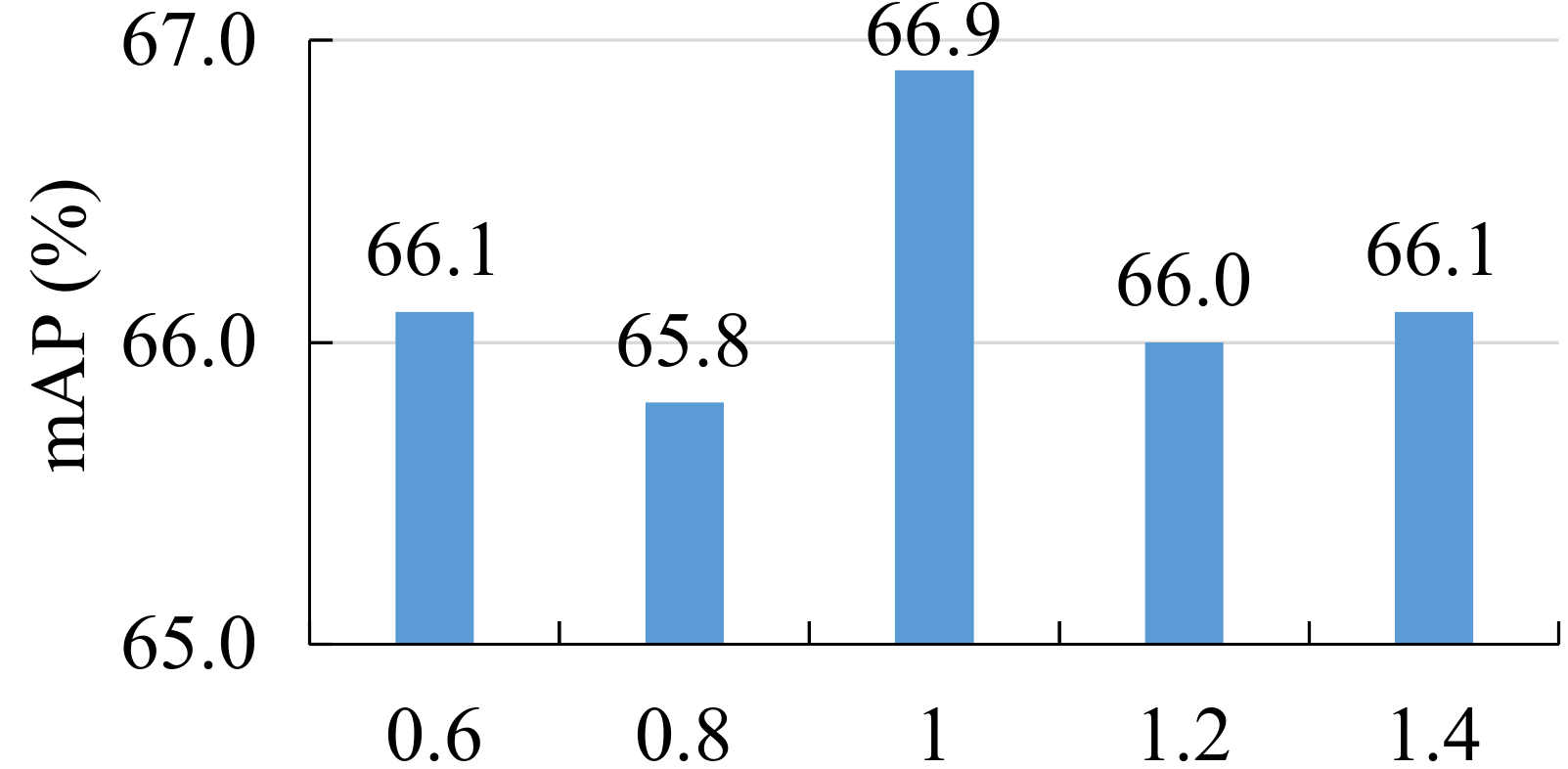}
         \caption{\footnotesize Influence of coefficient $\lambda$ in the loss function.}
         \label{figAblChannel}
     \end{subfigure}
     \begin{subfigure}[b]{0.22\textwidth}
         \centering
         \includegraphics[width=\textwidth]{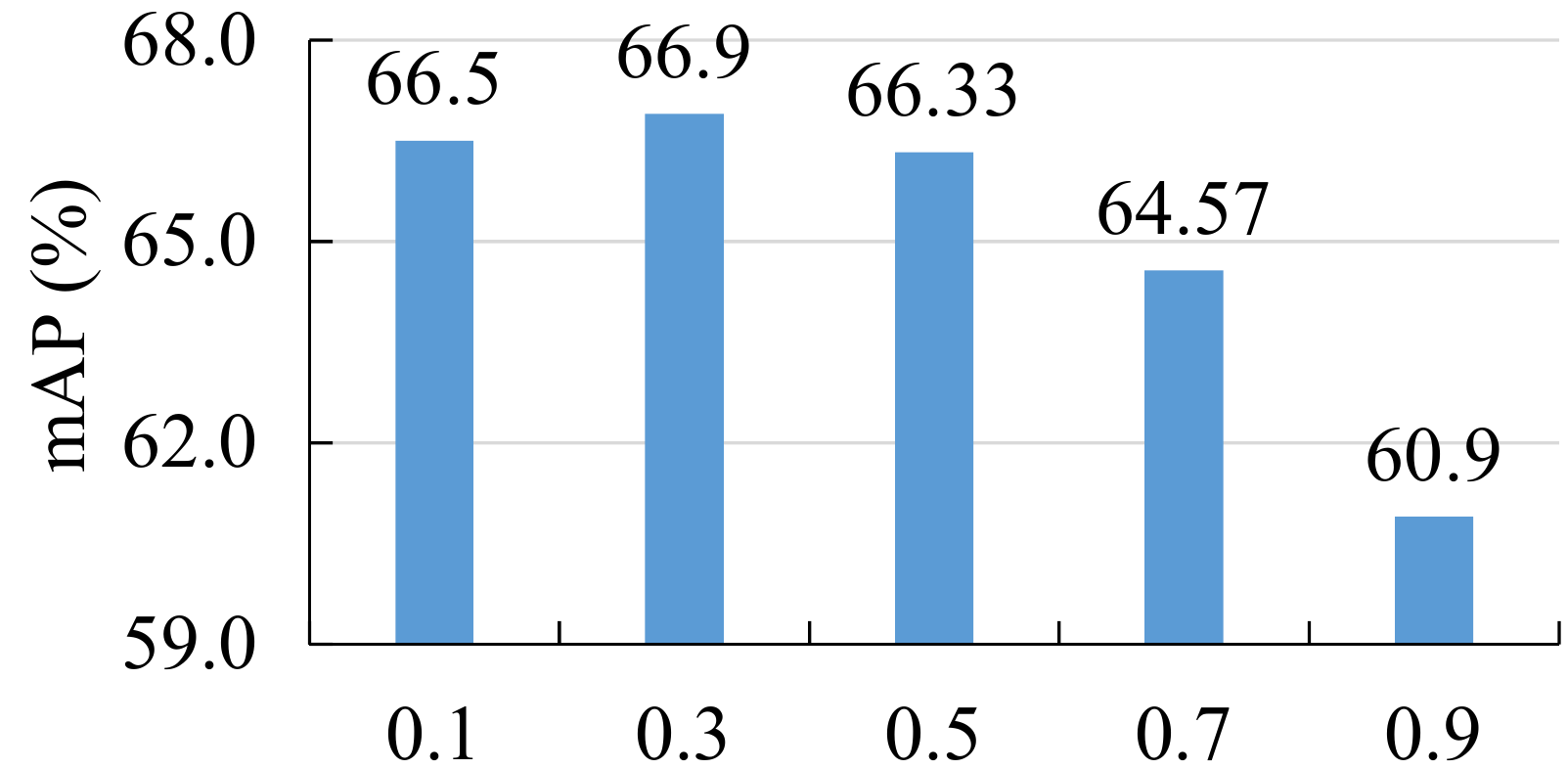}
         \caption{\footnotesize Influence of coefficient $\beta$ when fusing prediction scores.}
         \label{figAblSmt}
     \end{subfigure}
        \caption{Ablation studies about hyper-parameters in the proposed Colar method, measured by mAP (\%) on THUMOS14 dataset.}
        \label{figAblation}
    \vspace{-0.3cm}
\end{figure}

\subsection{Ablation experiments}

\textbf{Efficacy of each component.} The proposed Colar method consists of the dynamic branch and the static branch, as well as a consistency loss $\mathcal{L}_{cons}$ to enable mutual guidance between two branches. Table \ref{tabAblComplements} studies the efficacy of each component on all three benchmark datasets. Firstly, the dynamic branch performs superior to the static branch, demonstrating the necessity of carefully modeling temporal dependencies. Besides, without the consistency loss, directly fusing prediction scores of two branches (\eg using Eq. (\ref{eqFusion})) only observes limited improvements, while $\mathcal{L}_{cons}$ can further improve the detection performance.

\textbf{Ablations about the dynamic branch.} As shown in Figure \ref{figAblation} (a), we first study the influence of temporal scope $T$ in modeling temporal dependencies and find 64 is a proper choice for the dynamic branch. The too-short temporal scope is insufficient to perceive evolvement within a video segment, while too long temporal scope would bring noises. In addition, we vary the number of convolutional layers and choose two layers, as shown in Figure \ref{figAblation} (b).

\textbf{Ablations about the static branch.} Based on K-Means clustering, the number of exemplars is an influential parameter for the static branch. As shown in Figure \ref{figAblation} (c), the ability of limited exemplars is insufficient and overwhelming exemplars would damage the performance as well.

\textbf{Ablations about the complete method.} Given the complete method, Figure \ref{figAblation} (d) studies the performance under different feature channels and verifies 1024 is a proper choice. Figure \ref{figAblation} (e) studies the coefficient $\lambda$ for the consistency loss in the training phase, while Figure \ref{figAblation} (f) verifies the influence of coefficient $\beta$ in the inference phase. We find $\lambda=1$ and $\beta=0.3$ are proper choices.

\begin{figure}[thbp]
\centering
\includegraphics[width=1\linewidth]{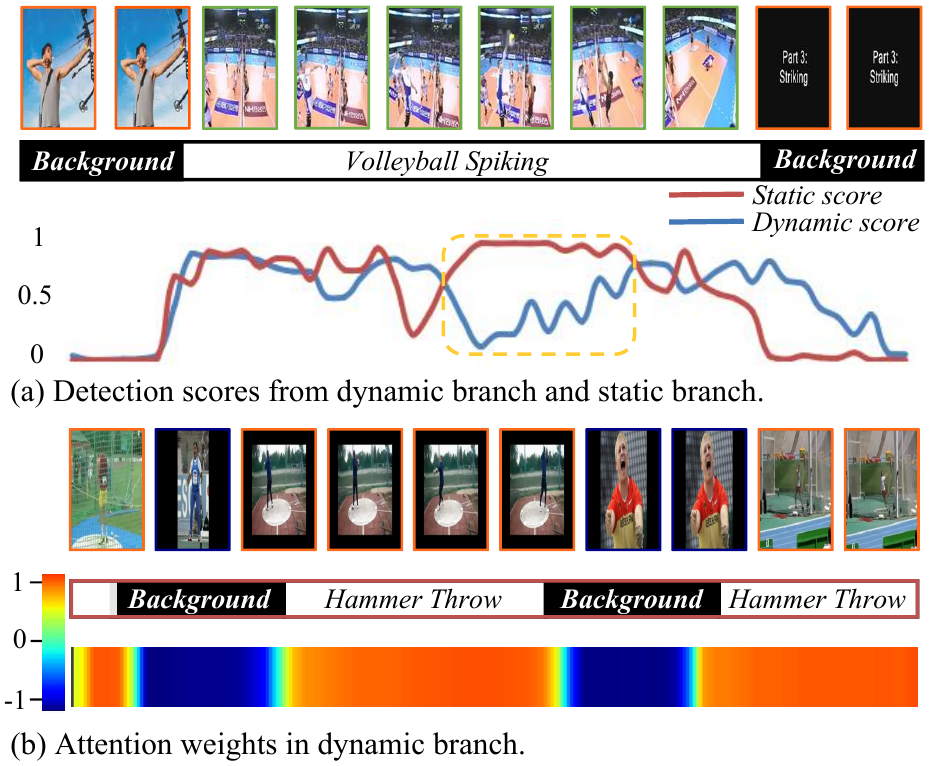}
\caption{Qualitative analysis of the proposed Colar method.}
\label{figVisualization}
\vspace{-0.3cm}
\end{figure}

\subsection{Qualitative analysis}
Figure \ref{figVisualization} qualitatively analyzes the proposed Colar method. Figure \ref{figVisualization} (a) exhibits the static and dynamic scores within a video segment. Because the \textit{Volleyball Spiking} instance shows dramatic viewpoint changes, the dynamic branch predicts low confident scores for some unique frames (shown in the yellow dotted box). In contrast, the static branch consults representative exemplars from the \textit{Volleyball Spiking} category and consistently predicts high scores for these unique frames. Figure \ref{figVisualization} (b) presents a video segment containing multiple action instances, exhibits the similarity between the current frame (the last one) and its historical frames. The similarity weights clearly highlight historical action frames and suppress background frames, which contributes to aggregating temporal features.

\section{Conclusion}
This paper proposes Colar, based on the exemplar-consultation mechanism, to conduct category-level modeling for each frame and capture long-term dependencies within a video segment. Colar compares a frame with exemplar frames, aggregates exemplar features, and carries out online action detection. In the dynamic branch, Colar regards historical frames as exemplars and models long-term dependency with a lightweight network structure. In the static branch, Colar employs representative exemplars of each category and captures the category particularity. The prominent efficacy of Colar would inspire future works to pay attention to category-level modeling. In addition, as Colar has made a good trade-off between effectiveness and efficiency, it is a promising direction to conduct online action detection directly from streaming video data, which can benefit practical usage.

\noindent \textbf{Limitations.} Because Colar is only verified on the benchmark datasets, it may observe performance drop in practical scene due to new challenges,~\eg long-tail distribution, open-set action categories. Besides, the unintended usage of Colar for surveillance may violate individual privacy.

\noindent
\textbf{Acknowledgments.} This work was supported in part by the Key-Area Research and Development Program of Guangdong Province (No.2019B010110001) and the National Natural Science Foundation of China under Grant U21B2048, 62036011, and the Open Research Projects of Zhejiang Lab (No.2019KD0AD01/010).

%%%%%%%%% REFERENCES
{\small
\bibliographystyle{ieee_fullname}
\bibliography{egbib}
}

\end{document}